\documentclass[conference]{IEEEtran}
\IEEEoverridecommandlockouts
\usepackage{cite}
\usepackage{amsmath,amssymb,amsfonts}
\usepackage{graphicx, color}
\usepackage{subcaption}
\usepackage[export]{adjustbox}
\usepackage{textcomp}
\usepackage{dblfloatfix}
\usepackage{url} 
\usepackage[table]{xcolor} 
\usepackage{paralist}
\usepackage{enumitem}
\usepackage{multirow}
\usepackage{hyperref}
\usepackage{algorithmic}
\usepackage{pifont}
\usepackage[linesnumbered,ruled,vlined]{algorithm2e}
\usepackage{booktabs}
\hypersetup{
    colorlinks=true,
    linkcolor=red, 
    citecolor=blue, 
    urlcolor=magenta    
}

\def\BibTeX{{\rm B\kern-.05em{\sc i\kern-.025em b}\kern-.08em
    T\kern-.1667em\lower.7ex\hbox{E}\kern-.125emX}}

\newcommand{\smallurl}[1]{\footnotesize\url{#1}}

\definecolor{baselinecolor}{gray}{.9}

\begin{document}

\title{Robust Fairness Vision-Language Learning for Medical Image Analysis}

\author{\IEEEauthorblockN{Sparsh Bansal\textsuperscript{1}, Mingyang Wu\textsuperscript{1},  Xin Wang\textsuperscript{2}, Shu Hu\textsuperscript{1}$^*$\thanks{$^*$Corresponding Author} }
\IEEEauthorblockA{
{\textsuperscript{1}Purdue University, West Lafayette, USA {\tt \small \{bansa125, wu2415,  hu968\}@purdue.edu} }\\
\textsuperscript{2}University at Albany, State University of New York, New York, USA {\tt \small xwang56@albany.edu}}
}

\maketitle
\thispagestyle{plain}
\pagestyle{plain}

\begin{abstract}
The advent of Vision-Language Models (VLMs) in medical image analysis has the potential to help process multimodal inputs and increase performance over traditional inference methods. However, when considering the domain in which these models will be implemented, fairness and robustness are important to ensure the model stays true for any patient. In this paper, we introduce a framework for ensuring robustness and fairness of VLM models. This framework modifies the loss function at training by identifying and adjusting faulty image-text pairs through a Dynamic Bad Pair Mining algorithm and also utilizing Sinkhorn distance to ensure the loss distributions of protected groups do not deviate from the total loss. Experimental testing of our framework shows up to a 8.6\% improvement when looking at equity-scaled AUC. The code is available at \smallurl{https://github.com/Purdue-M2/Robust_Fairness_for_Medical_Image.git}.

\end{abstract}


\section{Introduction}

Machine learning assisted detection has been increasing in interest and research due to the extended capabilities of ML models to interact with multiple modalities of data and parse thousands of data points quickly, as well as provide a base for interaction with the usage of LLMs \cite{chen2024survey,yang2024llm,lin2024robust,lin2024robust1,peng2024uncertainty,lin2024detecting}. These models provide a much more efficient and cost-effective method of analysis. However, when considering the domain in which these models are applied, the question of equality and equity come up. If these models have inherent biases due to dataset distributions or model training can lead to those biases being applied when diagnosing patients or informing medical experts, which can negatively impact certain groups and create ethical dilemmas. Therefore, ensuring that medical detection models are equitable and fair is a cornerstone of the research to ensure that patient healthcare is fair and socioeconomic issues are not exacerbated.

The problem of fairness \cite{wang2025fg,wang2025towards,wu2025preserving,lin2025ai,lin2024preserving,hu2024fairness,ju2024improving,hu2022distributionally} in machine learning models used has shown to be a issue. Reviewing previous research shows that 67\% of papers selected that covered applications of machine learning in the medical domain showed that a racial bias existed in their model \cite{huang2022evaluation}. However, with the advent of Vision-Language models that allow for multimodal input and output, racial biases also play a factor that aren't accounted for. A study into CLIP \cite{radford2021learning}, a popular pre-trained model used as the base for most VLM-based medical analysis models, shows that it contains inherent biases towards certain races, for example, associating words like  “psychopath”, “felon”, “gang-related”,“terrorist”, and “fraud”, with races like  white, black, Latino/Hispanic, Middle Eastern, and Indian respectively \cite{hamidieh2024identifying}. Therefore, accounting for racial biases in VLM models trained for the medical domain is a priority to ensure equity in the healthcare sector. However, the current domain focuses on general biases in pretrained models, which cannot account for biases when fine-tuning for specific medical applications. 

\begin{figure}[!tp]
    \includegraphics[width=0.5\textwidth]{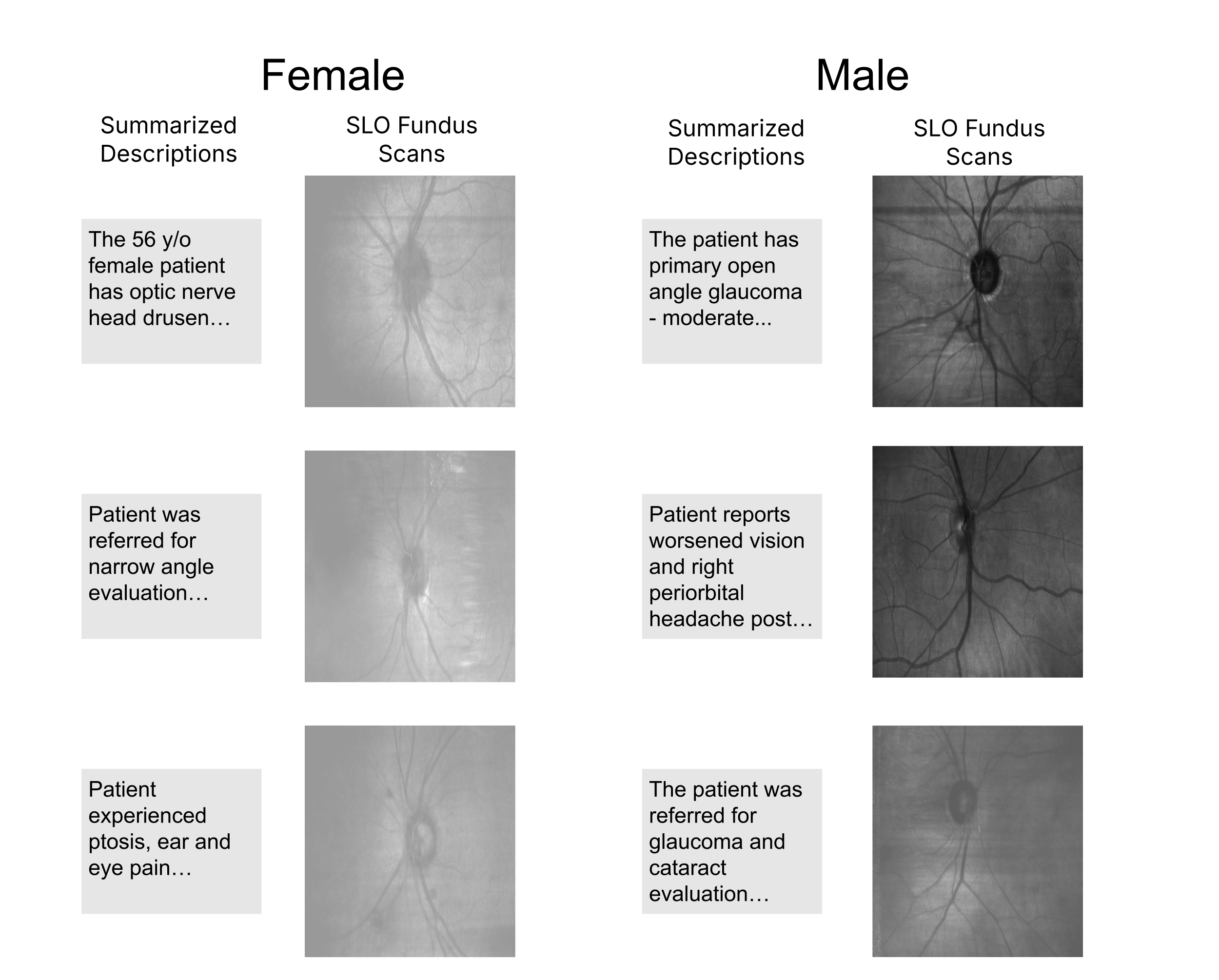}
    \vspace{-5mm}
    \caption{\textit{A few examples from the Harvard-FairVLMed database \cite{luo2024fairclip} that is used for training and fairness optimization, showing SLO Fundus photography images for glaucoma diagnosis.}}
    \vspace{-4mm}
    \label{fig:overview}
\end{figure}

\begin{figure*}[!tp]
    \centering
    \includegraphics[width=1\textwidth]{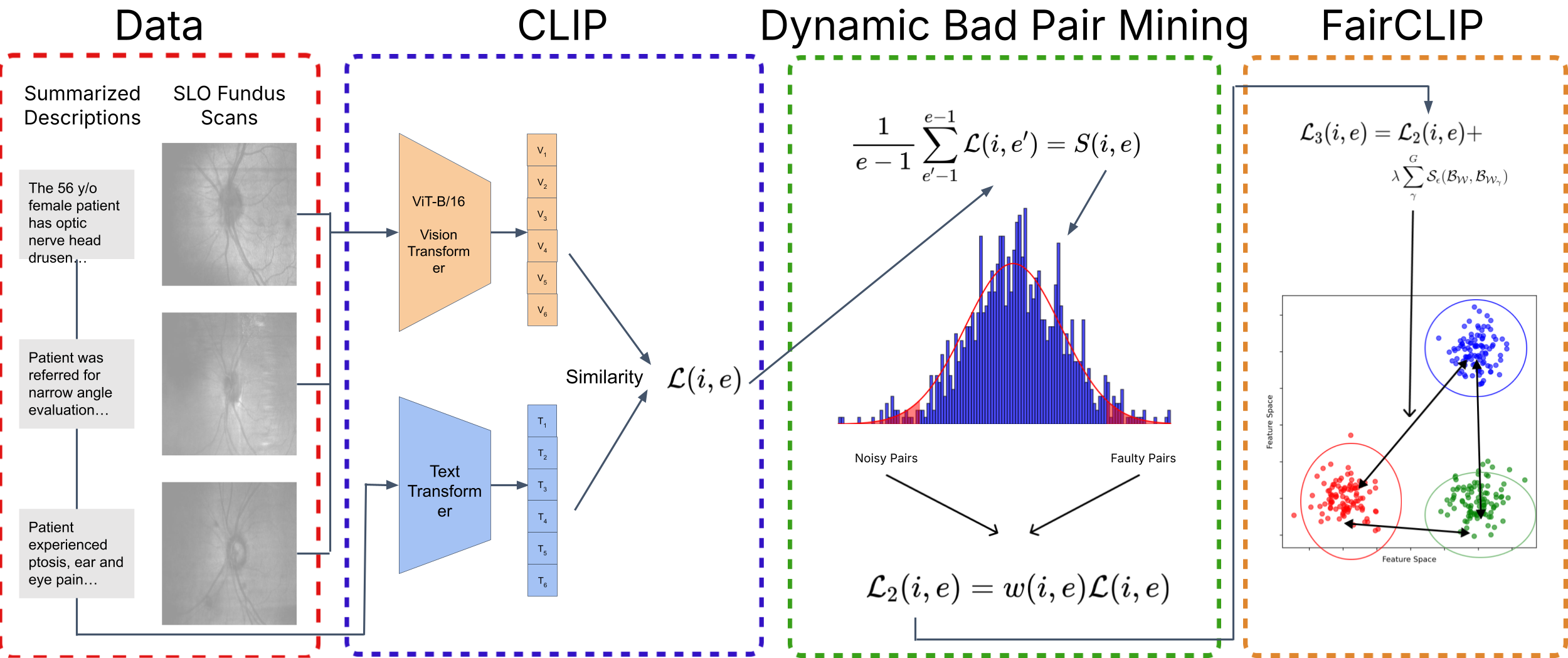}
    \vspace{-5mm}
    \caption{\textit{Overview of our proposed model using CLIP ViT-B/16 \cite{dosovitskiy2020image} as the base model with the loss function algorithms to account for erroneous pairs through Bad Pair Mining and equity through the Sinkhorn distance}}
    \vspace{-4mm}
    \label{fig:SL}
\end{figure*}

The robustness \cite{huang2024robustly,huang2025diffusion,tsai2024uu,yang2024explainable,lin2024robust, lin2024robustcovid, lin2024robustlight, wang2024u,sun2024repmedgraf,zhu2024cgd,wang2024u,hu2024umednerf,wang2024neural,hu2023attention,hu2023rank,hu2022pseudoprop,guo2022robust,hu2022rank,hu2022sum,hu2020learning} of a model is also a consideration when considering the application of the model in the healthcare sector. When testing the robustness of VLMs to data perturbations and modifications, it was found that not only are models less robust when both the text and image are faulty, but also that due to bias, models are less robust when data is either all-female or if the perturbations occur on these biased characteristics, such as swapping male to female \cite{schiappa2022robustness}. Therefore, improving the robustness of the model will improve its performance dealing with real-world data, especially erroneous data, and limit the performance impact due to perturbations on characteristics like race. This is especially important in the scope of medical VLM training. Medical image datasets usually contain long, specific medical expert notes or diagnosis, and this creates a problem for most modern VLMs. CLIP \cite{radford2021learning}, a popular VLM for medical image analysis, for example, can only take a maximum of 77 text tokens, but an empirical study shows that the effective length is closer to 20 tokens \cite{zhang2024long}. Therefore, modifications must be made, and generally these clinical notes are summarized through a general LLM such as GPT-4 \cite{openai2023gpt} or medical specific LLMs such as PMC-LLAMA \cite{wu2024pmc}. However, compressing data through the usage of LLMs can lead to noise or even faulty information in the textual data, propagating to errors in training and inference. Therefore, robustness is vital in designing VLMs for the medical domain as the size of medical datasets needed for training the model requires the model to be able to handle noise or faulty image-text pairs that may be caused from summarization. Being able to account for erroneous pairs will allow scalability to more datasets and better inference, allowing applications in the medical sector to be practical. 

Glaucoma is a problem that affects millions globally, but is a disease that has an unequal impact on minority groups, with the disease being most common in non-Hispanic blacks, then non-Hispanic whites, Mexican Americans, and others. Not only that, but it is a disease that is relatively unnoticed, as in the same study, they found that 55\% of the people diagnoses with glaucoma were unaware of the issue \cite{gupta2016prevalence}. Therefore, the application of robust, fair VLMs can improve the situation while accounting for equity. 

In this work, we propose a robust and fair VLM based framework as depicted in Figure \ref{fig:SL}. This framework consists of a VLM-based pretrained model, such as CLIP \cite{radford2021learning}, a optimal transport based fairness loss function \cite{luo2024fairclip}, and a Dynamic Bad Pair Mining algorithm to detect and account for erroneous pairs during training \cite{lan2023towards}. This framework can be used to train with image-text pairs to generate diagnosis labels, and can be applied to any feature based VLM, which is also tested by swapping for another VLM model architecture, BLIP-2 \cite{li2023blip}, and producing improved results.

Our contributions are summarized as follows:
\begin{enumerate}
    \item We propose the first robust and fair framework for VLM-based medical image detection and diagnosis
    \item Our method outperforms the pretrained baseline as well as a non-robust architecture
    \item Our method is scalable and is shown to be applicable to any feature based VLM, showing improvement independent of model architecture
\end{enumerate}

\section{Related Work}
\subsection{Vision-Language Models in Medical Image Analysis}

Much research has been done on applying Vision-Language Models (VLMs) to Medical Image Analysis. With the advent of multi-modal models such as LLaVA or CLIP, these VLMs have shown to be effective in analyzing low resolution data and being able to interact and follow up with their results, something CNNs and other methods cannot \cite{van2024large} \cite{he2023transformers}. In this domain, CLIP \cite{radford2021learning} has shown to be an effective and optimal method due to its multi-modal design and zero-shot capabilities, allowing training of text-image pairs that can make quick and optimal inferences. Papers such as \cite{kakkar2024language}, \cite{shi2024eyeclip}, and \cite{wu2024facmic} use the CLIP architecture to develop medical image analysis models.

\subsection{Robust VLMs in Medical Image Analysis}
Applying VLMs to the domain requires a certain robustness and replicability, as described by \cite{subedi2025reliability}. In the context of healthcare where a certain dependency is placed on the model, we must evaluate the consistency, accuracy, and robustness to training and test data to ensure their safe and consistent application. One such issue comes up with the data used train contrastive models such as CLIP \cite{radford2021learning}, where positive pairs can be faulty or erroneous. The importance of robustness especially comes by the fact that medical notes are often summarized due to their length and variability, since most VL models are most optimal on short text, unlike the lengthy clinical notes \cite{luo2024fairclip}. These LLM summarizations can cause errors or generate noise that can lead to faulty training. Catching and accounting for the noise or faults generated due to summarizations, especially when dealing with LLMs, is necessary as the dataset is expanded, with multiple sources of clinical notes and compounded errors. To this end, there exist many improvements to single modal large contrastive models to resist variations or errors in training datasets, such as generating stronger augmentations \cite{wang2022contrastive}, arbitrary pair learning \cite{wu2024rethinking}, or Dynamic Bad Pair Mining (DBPM) \cite{lan2023towards}. However, all these methods focus on single modal applications such as images or time series data. In this paper, we focus on applying Dymanic Bad Pair Mining \cite{lan2023towards} in a multimodal context to increase the robustness of a image-text based contrastive model. Along with this, robust methods do not consider that correct pairs and data may be biased towards certain demographics, leading to an implicit bias in the model, especially in medical image analysis where the effects of bias are visible


\subsection{Fair VLMs in Medical Image Analysis}
An important ethical consideration for any model applied in the medical domain is the fairness of the model to be effective for any demographic. According to \cite{chingnabe2025vision}, ethical considerations such as the equal application and performance of these models on diverse populations is necessary to ensure disparities in healthcare access are not widened. To this end, multiple solutions have been designed to help normalize the implicit biases caused by imperfect training datasets \cite{xu2023fairness}. FairCLIP \cite{luo2024fairclip} is one such model that aims to account for fairness in a VLM domain, using a carefully curated dataset for glaucoma fundus images and a fairness algorithm implemented during training. However, robustness to variations in training data pairs and repeated training are overlooked, focusing more on the fairness of the model. To that end, our paper aims to build on this model to ensure robustness and repeatability.  

\begin{table*}[t]
\centering
\begin{tabular}{@{}cllllllll@{}}
\toprule
\textbf{Attribute}                                      & \multicolumn{1}{c}{\textbf{Model}} & \textbf{DPD} $\downarrow$ & \textbf{DEOdds} $\downarrow$ & \textbf{AUC} $\uparrow$ & \textbf{ES-AUC} $\uparrow$ & \multicolumn{3}{c}{\textbf{Group-wise AUC} $\uparrow$}                \\ \midrule
\multirow{4}{*}{\textbf{Race}}                          & \multicolumn{1}{c}{}               &                &                 &                &                 & \textbf{Asian}        & \textbf{Black}    & \textbf{White} \\
                                                        & CLIP\cite{radford2021learning}                               & 12.86          & 20.79           & 63.68          & 56.76           & 66.44                 & 70.55             & 61.12          \\
                                                        & FairCLIP\cite{luo2024fairclip}                           & 14.1           & 17.8            & 67.6           & 65.7            & 65.6                  & 68.1              & 67.2           \\
                                                        & Robust FairCLIP                    & \textbf{11.05} & \textbf{11.16}  & \textbf{70.84} & \textbf{65.88}  & \textbf{75.16}        & \textbf{72.54}    & \textbf{69.34} \\ \midrule
\multirow{4}{*}{\textbf{Gender}}                        &                                    &                &                 &                &                 & \textbf{Female}       & \textbf{Male}     &                \\
                                                        & CLIP  \cite{radford2021learning}                             & 2.52           & \textbf{3.34}   & 63.68          & 59.93           & 61.02                 & 67.28             &                \\
                                                        & FairCLIP \cite{luo2024fairclip}                          & \textbf{0.35}  & 5.54            & 67.6           & 64.7            & 65.6                  & 70.1              &                \\
                                                        & Robust FairCLIP                    & 2.71           & 7.08            & \textbf{70.84} & \textbf{66.37}  & \textbf{67.89}        & \textbf{74.63}    &                \\ \midrule
\multicolumn{1}{l}{\multirow{4}{*}{\textbf{Ethnicity}}} &                                    &                &                 &                &                 & \textbf{Non-Hispanic} & \textbf{Hispanic} &                \\
\multicolumn{1}{l}{}                                    & CLIP  \cite{radford2021learning}                             & \textbf{4.45}  & \textbf{4.98}   & 63.68          & 59.31           & 63.93                 & 56.56             &                \\
\multicolumn{1}{l}{}                                    & FairCLIP \cite{luo2024fairclip}                          & 5.35           & 8.88            & 67.6           & \textbf{64.1}   & 67.8                  & 62.3              &                \\
\multicolumn{1}{l}{}                                    & Robust FairCLIP                    & 9.85           & 14              & \textbf{70.84} & 61.71           & \textbf{71.35}        & \textbf{56.56}             &                \\ \midrule
\multicolumn{1}{l}{\multirow{4}{*}{\textbf{Language}}}  &                                    &                &                 &                &                 & \textbf{English}      & \textbf{Spanish}  & \textbf{Other} \\
\multicolumn{1}{l}{}                                    & CLIP  \cite{radford2021learning}                             & \textbf{11.73} & \textbf{13.3}   & 63.68          & 57.52           & 63.76                 & 60.8              & 55.94          \\
\multicolumn{1}{l}{}                                    & FairCLIP \cite{luo2024fairclip}                          & 18.1           & 16.3            & 67.6           & \textbf{59.5}   & 67.8                  & \textbf{70.2}     & 56.6           \\
\multicolumn{1}{l}{}                                    & Robust FairCLIP                    & 12.92          & 23.77           & \textbf{70.84} & 58.59           & \textbf{71.2}         & 61.36             & \textbf{59.77} \\ \bottomrule
\end{tabular}
\caption{This summarizes the fairness metrics computed for the two baseline models, CLIP \cite{radford2021learning} and FairCLIP \cite{luo2024fairclip}, along with our RobustFairCLIP model. For all the values, the arrow indicated whether a value should be less or more to indicate better performance. Bolded values represent the best value in the group. }
\label{tab:results}
\end{table*}

\section{Method} 
\subsection{Overview}
 Figure \ref{fig:SL} provides an overview of the model structure. The one-shot classifier is based off the ViT-B/16 CLIP architecture \cite{dosovitskiy2020image} \cite{radford2021learning}. To account for fairness, differences in distributions of demographics are accounted in the loss \cite{luo2024fairclip}. To improve robustness, the Dynamic Bad Pair Mining algorithm \cite{lan2023towards} is used on the loss to account for variations in the dataset.

\subsection{One-shot Contrastive Learning with CLIP}
With dataset $\mathcal{D} = \{x_{i}, y_i, d_i, p_i \}$ where $x_i \in X$ represents a Scanning Laser Ophthalmoscopy (SLO) fundus image, $y_i \in Y$ represents the paired GPT-4 summarized medical text note for the image, $d_i \in D$ represents the diagnosis for glaucoma for the image, and $\gamma_i \in G$ represents the protected attributes and demographics, such as race, for the associated image. Using the CLIP model \cite{radford2021learning}, we train an image and text encoder to generate normalized image and text features $f_X$ and $f_Y$. Given a batch $B$, The optimization aims to maximize the cosine similarity of the $B$ positive pairs while minimizing the cosine similarity of the $B^2-B$ false pairs. This is done by generating a similarity matrix $M \in \mathbb{R}^{N \times N}$ where $M = f_{Xi}^{\top}\cdot f_{Yi}$. From \cite{luo2024fairclip}, the optimization goal is then
\begin{align*}
    \min_f \sum \limits^{N}_{i=1} \sum \limits^{N}_{j=1} \delta(i-j)\log \left( \frac{f_{Xi}^{\top}\cdot f_{Yi}}{||f_{Xi}||\cdot ||f_{Yi}||}\right)
\end{align*}
with $\delta(i-j)$ as Dirac's Delta function with the modification that $\delta(0) = 1$. To get a numerical loss value for gradient optimization in a batch, we modify the optimization goal. From \cite{radford2021learning}, given similarity matrix $W$ computed in batch $\mathcal{W}$, we can apply symmetric cross entropy loss \cite{wang2019symmetric} as follows
\begin{align}\label{eq:loss1}
    \mathcal{L}_1 = \left( - \sum\limits^N_{i=1} d_i\log(\text{row}_i(W)) - \sum\limits^N_{i=1} d_i\log(\text{col}_i(W))\right)/2
\end{align}

\subsection{Dynamic Bad Pair Mining}
The summarization of clinical notes used in training can lead to noisy or even faulty image-text pairs. This will cause erroneous models, and therefore, to improve robustness, we employ a Dynamic Bad Pair Mining (DBPM) algorithm \cite{lan2023towards}. This algorithm, modified from a single-modal context, allows us to identify noisy or faulty pairs at training and adjust their loss contribution, accounting for the noise. 

We begin with the loss function in equation \eqref{eq:loss1}, and record the average historical loss $S(i,e)$ for each batch $i$ in epoch $e$, given by this function
\begin{align*}
    S(i,e) = \frac{1}{e-1} \sum\limits^{e-1}_{e'=1} \mathcal{L}_{1_{(i,e)}}
\end{align*}
where $\mathcal{L}_{1_{(i,e)}}$ is the loss as epoch $e$, batch $i$. We then generate a statistical description, the mean and standard deviation, of the training loss history of the $e$-th epoch assuming it follows a Gaussian distribution, giving us
\begin{align*}
    \mu_e = \frac{1}{N}\sum\limits^N_{i =1} S(i,e)& &\sigma_e = \sqrt{\frac{\sum\limits^N_{i=1} (S(i,e) - \mu_e)^2}{N}}
\end{align*}
We threshold the noisy and faulty pairs by assigning the tails as faulty. We choose hyperparameters $\alpha$ and $\beta$ to determine the percentile of batches that are considered noisy or faulty, with the thresholds as
\begin{align}\label{eq:hyperparams}
    a = \mu_e - \alpha \sigma_e & & b = \mu_e + \beta \sigma_e
\end{align}
where the batches at epoch $e$ are considered to be correct pairs if the loss falls within $[a,b]$, noisy batches if the loss falls within $(\infty, a)$ and faulty batches if the loss falls within $(b,\infty)$. For noisy or faulty pairs, we use a smooth weight reduction as described by \cite{lan2023towards}, given by this equation
\begin{align*}
    \mathbf{R}(\mathcal{L}_{1_{(i,e)}}, \mu_e, \sigma_e) = \frac{1}{\sigma_e \cdot \sqrt{2 \pi}} \exp \left(-\frac{(\mathcal{L}_{1_{(i,e)}} - \mu_e)^2}{2\sigma^2_e} \right)
\end{align*}
Therefore, we can combine the thresholds and weight reduction to give us a reduction function given as
\begin{align*}
    w(i,e) = \begin{cases}
        1 &\text{if } \mathcal{L}_{1_{(i,e)}} \in [a, b] \\
        \mathbf{R}(\mathcal{L}_{1_{(i,e)}}, \mu_e, \sigma_e) &\text{if } \mathcal{L}_{1_{(i,e)}} \in (\infty, a) \cup (b,\infty)
    \end{cases}
\end{align*}
With the loss weight function, we can generate the weighted loss for each batch by weighting equation \eqref{eq:loss1}, given as follows
\begin{align}\label{eq:loss2}
    \mathcal{L}_2 = w(i,e)\mathcal{L}_1
\end{align}


\subsection{Fair Learning}
Inspired by \cite{luo2024fairclip}, we account for differences in the similarity distributions of each protected attribute $p_i \in P$. To achieve this, we take the distribution $\mathcal{B}(x, y, g|\theta)$, which represents the distribution of the positive pairs $M_{i,i}$, where $\theta$ represents the model instance and $g$ represents attributes. Therefore, the optimization problem can be represented by
\begin{align}\label{eq:sinkhornopt}
    \min_\theta \sum\limits^G_\gamma d(\mathcal{B}(x, y, g|\theta) - \mathcal{B}(x, y, g|g=\gamma, \theta)
\end{align}
with $d$ being some distance function. To make this computationally plausible, we calculate discrete distributions for each batch $\mathcal{W}$, where $\mathcal{B}_{\mathcal{W}}$ and $\mathcal{B}_{\mathcal{W}_\gamma}$ are the distributions with $\mathcal{B}_{\mathcal{W}_\gamma}$ being the distribution of attribute $\gamma$ in batch $\mathcal{W}$. We use Sinkhorn distance \cite{luo2024fairclip} \cite{peyre2019computational} as our distance function, given by

\begin{align}\label{eq:sinkhorn}
    \mathcal{S}_\epsilon(\mathcal{B}_{\mathcal{W}}, \mathcal{B}_{\mathcal{W}_\gamma}) = \min_{\mathbf{P} \in \mathbf{U}(\mathcal{B}_{\mathcal{W}}, \mathcal{B}_{\mathcal{W}_\gamma})} \langle \mathbf{P}, \mathbf{C} \rangle - \epsilon \mathbf{H}(\mathbf{P} | \mu \otimes \nu )
\end{align}

where $\mathbf{U}(\mathcal{B}_{\mathcal{W}}, \mathcal{B}_{\mathcal{W}_\gamma})$ represents the joint distributions of $\mathcal{B}_{\mathcal{W}}$ and $\mathcal{B}_{\mathcal{W}_\gamma}$, $\langle \mathbf{P}, \mathbf{C} \rangle$ represents the estimated cost of going from $\mathcal{B}_{\mathcal{W}}$ to $\mathcal{B}_{\mathcal{W}_\gamma}$ with cost function $\mathbf{C}$, and $\mathbf{H}(\mathbf{P} | \mu \otimes \nu ) $ represents the the relative entropy function with respect to the probabilistic distributions $\mu$ and $\nu$, with a parameter $\epsilon$ to control the strength of the entropy, known as "blur". Combining the optimization problem \eqref{eq:sinkhornopt} with equation \eqref{eq:sinkhorn} and adding the loss to equation \eqref{eq:loss2} gives us our final loss function
\begin{align}\label{eq:loss3}
    \mathcal{L}_3 = \mathcal{L}_2 + \lambda \sum\limits^G_\gamma \mathcal{S}_\epsilon(\mathcal{B}_{\mathcal{W}}, \mathcal{B}_{\mathcal{W}_\gamma})
\end{align}

where $\lambda$ is a regularization parameter for the sinkhorn loss. 

\subsection{Pre-trained Models}
The pre-trained model used was the OpenAI CLIP ViT-B/16 model \cite{radford2021learning} \cite{dosovitskiy2020image}. CLIP is a multi-model transformer based model developed by OpenAI. It has efficient zero-shot transfer capabilities. This model leverages 2 transformers to generate text and visual features that it trains to maximize similarity on \cite{radford2021learning}. The vision transformer used in this pretrained model is the ViT-B/16, which is an efficient alternative to ResNet, along with using a patch size of 16x16 \cite{dosovitskiy2020image}. The CLIP ViT-B/16 was trained on 400 million image-text pairs to form the basis of the model we used. 
The CLIP ViT-B/16 was chosen for its efficiency and multi-modal transformer design \cite{luo2024fairclip}. The design of image-text pair similarity allow the model to be finetuned on the specific features of the glaucoma diagnosis, along with being able to use the generated features to account for divergence in fairness and performance between protected groups. Utilizing transfer learning allows us to adjust the transformers to account for variations and demographic differences. 

To test improvement across model architectures, we also tested our improvements on another VLM pre-trained model architecture, BLIP-2 \cite{li2023blip}. This is another multi-modal VLM that leverages a Querying Transformer that allows greater efficiency and performance with less parameters. BLIP-2 uses CLIP ViT-L/14 as the frozen vision transformer. By bootstrapping frozen image and text LLMs, they're able to efficiency perform zero-shot multimodal tasks. 

\begin{table*}[!htbp]
\centering
\begin{tabular}{@{}cllllllll@{}}
\toprule
\textbf{Attribute} &
  \multicolumn{1}{c}{\textbf{Model}} &
  \textbf{DPD} $\downarrow$ &
  \textbf{DEOdds} $\downarrow$ & 
  \textbf{AUC} $\uparrow$ &
  \textbf{ES-AUC} $\uparrow$ &
  \multicolumn{3}{c}{\textbf{Group-wise AUC} $\uparrow$} \\ \midrule
\multirow{3}{*}{\textbf{Race}} &
   &
   &
   &
   &
   &
  \textbf{Asian} &
  \textbf{Black} &
  \textbf{White} \\
 &
  BLIP-2 \cite{li2023blip}&
  \textbf{4.81} &
  12.96 &
  80.03 &
  77.77 &
  81.95 &
  79.13 &
  80.11 \\
 &
  Robust FairBLIP-2 &
  6.95 &
  \textbf{10.54} &
  \textbf{81.96} &
  \textbf{78.05} &
  \textbf{85.32} &
  \textbf{80.34} &
  \textbf{81.98} \\ \midrule
\multirow{3}{*}{\textbf{Gender}} &
   &
   &
   &
   &
   &
  \textbf{Female} &
  \textbf{Male} &
   \\
 &
  BLIP-2\cite{li2023blip} &
  \textbf{0.01} &
  \textbf{6.85} &
  80.03 &
  75.25 &
  77.09 &
  83.43 &
   \\
 &
  Robust FairBLIP-2 &
  2.37 &
  9.29 &
  \textbf{81.96} &
  \textbf{76.69} &
  \textbf{78.75} &
  \textbf{85.62} &
   \\ \midrule
\multirow{3}{*}{\textbf{Ethnicity}} &
  \textbf{} &
   &
   &
   &
   &
  \textbf{Non-Hispanic} &
  \textbf{Hispanic} &
   \\
 &
  BLIP-2\cite{li2023blip} &
  14.35 &
  29.29 &
  80.03 &
  68.11 &
  80.7 &
  63.19 &
   \\
 &
  Robust FairBLIP-2 &
  \textbf{13.36} &
  \textbf{15.4} &
  \textbf{81.96} &
  \textbf{73.26} &
  \textbf{82.42} &
  \textbf{70.56} &
   \\ \midrule
\multicolumn{1}{l}{\multirow{3}{*}{\textbf{Language}}} &
  \textbf{} &
   &
   &
   &
   &
  \textbf{English} &
  \textbf{Spanish} &
  \textbf{Other} \\
\multicolumn{1}{l}{} &
  BLIP-2\cite{li2023blip} &
  \textbf{13.43} &
  \textbf{17.66} &
  80.03 &
  56.92 &
  80.93 &
  58.52 &
  61.84 \\
\multicolumn{1}{l}{} &
  Robust FairBLIP-2 &
  14.65 &
  18.34 &
  \textbf{81.96} &
  \textbf{71.94} &
  \textbf{82.43} &
  \textbf{73.86} &
  \textbf{76.6} \\ \bottomrule
\end{tabular}
\caption{This table summarizes the fairness metrics for the baseline model, BLIP-2 \cite{li2023blip}, and our RobustFairBLIP-2 model. For all the values, the arrow indicates whether a value should be less or more to indicate better performance. Bolded values represent the best value in the group}
\label{tab:my-table3}
\end{table*}



\section{Experiments}
\subsection{Experimental Settings}

\subsubsection{Dataset}
We use the training, validation, and test datasets of the Harvard-FairVLMed dataset \cite{luo2024fairclip}. This dataset contains 10,000 samples from 10,000 patients, each with a patient's SLO Fundus scan and its respective de-identified clinical notes. These notes are also included summarized with GPT-4 for more consistent length. Along with that, each image-text pair contains 6 demographic identity attributes such as race, ethnicity, marital status, preferred language, gender, and age. The dataset includes 819 Asian, 1,491 Black, and 7,690 White individuals. Females make up 56.3\% of the sample. Ethnically, 90.6\% are Non-Hispanic, 4.0\% Hispanic, and 5.4\% unspecified. Most prefer English (92.5\%), with few preferring Spanish (1.7\%) or other languages (0.8\%). Regarding marital status, 57.4\% are married or partnered, 26.4\% single, and the rest are divorced, separated, widowed, or unspecified. This dataset allows for image-text pair training as well as demographic analysis for each batch or data point. This set is divided into 7,000 training data points, 1,000 validation, and 2,000 test.

\subsubsection{Evaluation Metrics}
To show the performance metrics in terms of the performances between demographic groups that this model accounts for, we calculate the Equity Scaled Area Under the Curve (AUC) \cite{luo2024harvard} metric for each demographic group in the test dataset. This consists of taking the AUC over the test set of a specific demographic group and then normalizing that with the overall AUC. This allows us to see performance of the model for all the protected groups we are optimizing in order to ensure equality, and be able to see if groups have similar groupwise AUC scores. Along with that and the overall AUC, these metrics measure the performance of the models overall and for each demographic group.  

\subsubsection{Baseline Methods}
This robust model is tested in comparison to the pretrained base CLIP ViT-B/16 \cite{radford2021learning} and the FairCLIP \cite{luo2024fairclip} model with the same shared hyperparameters for all tested models.

\subsubsection{Implementation Details}
We use a pretrained CLIP ViT-B/16 model \cite{radford2021learning}. The Adam optimizer is employed with a learning rate of $1e-5$, betas of $(0.1, 0.1)$, and a weight decay of $6e-5$. The model is trained for 10 epochs with a batch size of 32. The hyperparameter $\epsilon=1e-4$ is the "blur" of the Sinkhorn loss from equation \eqref{eq:sinkhorn}. The hyperparameter $\lambda=1e-7$ is used to regulate the loss from equation \eqref{eq:loss3}. For the robust model, we set hyperparameters $\alpha$ and $\beta$ from equation \eqref{eq:hyperparams} to both be 3. These experiments were designed using PyTorch and trained on an NVIDIA A100 GPU.

\subsection{Results}

Our results for the CLIP architecture are summarized in Table \ref{tab:results}. It is clear that the RobustFairCLIP model outperforms the baseline FairCLIP and CLIP models in most categories. The robust model shows an overall \textbf{3\%} improvement over the FairCLIP baseline and a \textbf{8.6\%} improvement over the CLIP baseline over the total AUC scores, which is what the model optimizes. This shows that our model is performing better and able to generate more robust and accurate results across all protected attributes. 

Looking at an alternative VLM architecture, BLIP-2 \cite{li2023blip}, our results are summarized in Table \ref{tab:my-table3}. In comparison to the baseline BLIP-2 pretrained model, the robust model outperforms in almost every category. The robust model showed a \textbf{6.4\%} improvement over the baseline when considering total AUC, which is what the model aims to optimize. 





\section{Conclusion}
Applying Vision-Language Models (VLMS) to the medical domain for diagnosis or assistance can improve the quality and access to diagnosis for conditions such as glaucoma. However, VLM models do not currently account for bias or robustness, leading to biased models that can impact the implementation and fairness of these models in the healthcare sector. To address these issues, we developed a framework that accounts for fairness and robustness, applying a Dynamic Bad Pair Mining algorithm \cite{lan2023towards} to account for faulty training pairs and utilizing Sinkhorn loss over the batch distributions of protected attributes \cite{luo2024fairclip} to any VLM. This system allows the VLM to not only account for protected attributes but ensure that bigger datasets can be used and still ensure robustness by accounting for faulty pairs. Our experimental results confirm our model's better performance over the baseline.

\textbf{Limitation.} One limitation of the model is that the fairness loss only considers the difference between one protected attribute, such as race or gender. While the addition of this loss does improve the other protected attributes, they are not directly being optimized. This could mean that better performance could be achieved by accounting for all distributions of protected attributes.

\textbf{Future Work.} We plan to look at including multiple protected attributes as part of the Sinkhorn loss and applying our framework to different model architectures to test the generalizability across architectures. 

\smallskip
\smallskip
\noindent\textbf{Acknowledgments.} This work is supported by the U.S. National Science Foundation (NSF) under grant IIS-2434967 and the National Artificial Intelligence Research Resource (NAIRR) Pilot and TACC Lonestar6.
The views, opinions and/or findings expressed are those of the author and should not be interpreted as representing the official views or policies of NSF and NAIRR Pilot.

\bibliographystyle{ieeetr}
\bibliography{main}

\end{document}